# Blind Universal Bayesian Image Denoising With Gaussian Noise Level Learning

Majed El Helou ⬡, *Student Member, IEEE*, and Sabine Süsstrunk ⬡, *Fellow, IEEE*

*Abstract*—**Blind and universal image denoising consists of using a unique model that denoises images with any level of noise. It is especially practical as noise levels do not need to be known when the model is developed or at test time. We propose a theoretically-grounded blind and universal deep learning image denoiser for additive Gaussian noise removal. Our network is based on an optimal denoising solution, which we call fusion denoising. It is derived theoretically with a Gaussian image prior assumption. Synthetic experiments show our network's generalization strength to unseen additive noise levels. We also adapt the fusion denoising network architecture for image denoising on real images. Our approach improves real-world additive image denoising PSNR results for training noise levels and further on noise levels not seen during training. It also improves state-of-the-art color image denoising performance on every single noise level, by an average of $0.1 dB$, whether trained on or not.**

## I. INTRODUCTION

**I**MAGE denoising is a fundamental image restoration task applied in all image processing pipelines. An image denoiser can also be part of deep network models to improve the training of high-level vision tasks [27]. However, being an ill-posed inverse problem, denoising is challenging [14].

After the development of the best analytical solution, BM3D [8], [18], little improvement in denoising performance was achieved until the advent of deep learning denoisers [59]. Recent Convolutional Neural Network (CNN) based methods achieve state-of-the-art image denoising performance and are even faster than traditional optimization-based approaches [55]. The increased capacity of deep CNN models also addresses the limitation of previous multi-layer perceptron methods when it comes to denoising different levels of noise [5]. Well-designed CNN architectures can also outperform adversarial training methods in image restoration tasks [45].

Neural networks can be deep and wide and thus have large capacity to model complex functions [56], [61], by leveraging network regularization or normalization [21] and residual learning [19]. However, the complex functions modeled by the networks are not interpretable and have little connection to stochastic denoising. This is a limitation for training general models for denoising different noise levels. Denoisers are *blind* when they require no information about the noise level at test time, and *universal* when a single model can handle all noise levels. Blind universal models are important since knowing the noise level, at test time or ahead of training, is not a practical scenario for most applications.

We first mathematically derive a blind and universal denoising function under the theoretical assumption that the image prior is Gaussian. Our denoising function, which is optimal in stochastic expectation, is referred to as fusion denoising because it fuses the input with a prior weighted using the signal-to-noise ratio. It is optimized for additive Gaussian noise removal. Our experimental results show that the state-of-the-art denoiser DnCNN [59] can model an optimal fusion denoising function. However, it only models it for noise levels that are seen by the network during training. For unseen levels, our synthetic experiment's fusion network, called *Fusion Net*, far outperforms DnCNN. We show on synthetic data our improved generalization results.

The assumption that the image prior is Gaussian does not necessarily apply to real-world images. Building on the foundations of our theoretical solution, we adapt our *Fusion Net* by designing a second network that *learns* a fusion function for additive Gaussian noise removal. We call this new network *Blind Universal Image Fusion Denoiser (BUIFD)*. BUIFD improves state-of-the-art denoising performance on noise levels seen in training for grayscale and color images on the standard Berkeley test sets (BSD68 and CBSD68) [41]. Furthermore, we show that our generalization results to unseen noise levels obtained in our synthetic experiment extend to the denoising of the grayscale BSD68 test set. Indeed, the denoising performance on noise levels not trained on improves by multiple PSNR points. We present an extended denoising evaluation that covers other test datasets and other traditional and learning-based denoising methods.

Our main contributions are: (1) we theoretically derive an optimal fusion denoising function and integrate it into a deep learning architecture (Fusion Net) to evaluate the optimality of deep networks on a theoretical additive Gaussian noise removal task with known prior, (2) we show on synthetic data that the integration of the auxiliary fusion loss into our Fusion Net improves the network's generalization power bringing closer to the optimal solution, and (3) we develop a blind universal image fusion denoiser (BUIFD) network adapted to real images, and show that it outperforms the state of the art for Gaussian noise removal on multiple standard image processing test sets.

The paper is organized as follows. After a review of related work, we first lay the ground for our theoretical experiments. Our experiment allows us to assess the optimality of the networks on training noise levels and the generalization of trained networks to unseen Gaussian noise levels, in comparison to the optimal Bayesian solution. We then extend the Bayesian framework solution into our network designed for real images (BUIFD) whose exact prior is unknown to improve generalization. Experimental results on standard denoising benchmarks show that our denoising network outperforms the state of the art, especially on unseen noise levels.

## II. Related Work

Image denoising approaches in the literature can be divided into classical methods and the more recent deep-learning-based methods. One common aspect is, however, the leveraging of image priors to improve denoising results. For practical reasons, it is important for a denoiser to be blind and universal since the noise levels in noisy images might not be constant or known.

*Image Priors:* Whether they are in the form of assumptions made on image gradients [23], [35], [42], [51], sparsity [10], [15], self-similarity within images [4], [11], [53], hybrid approaches [30], or neural network weights given a certain architecture [3], [59], image priors are essential for denoising. Even traditional methods based on diffusion or filtering (in space [37] or in other domains [44]) rely on some priors. They, in all their forms and for multiple image restoration problems, can be discovered and tested heuristically [13], [23], learned with dictionaries [15], with Markov random fields [41], or with deep neural networks [59]. In our network, the prior takes the explicit form of learned feature representations.

*Noise Modeling:* Additive white Gaussian noise is not necessarily the best model in practical scenarios such as denoising raw images [3]. Nevertheless, a large part of the image denoising literature focuses on Gaussian denoising since it remains a fundamental problem. Images with noise following different, potentially data-dependent, distributions can be transformed into images with Gaussian noise, and transformed back [31], [38]. In addition, a Gaussian denoising solution can serve as a proximal [26], [36] for image regularizers. It can be a substitute for the costly step in half-quadratic splitting (HQS) optimization, typically responsible for non-differentiable regularization in image processing. This approach is taken in the recent HQS method that leverages the denoiser for image restoration [60]. We thus work with the assumption of an additive white Gaussian noise model.

*Image Denoisers:* Having to know the exact noise level is a serious limitation in practice for denoisers, and to know it ahead of time, before training, is even more limiting. A fixed and known noise level is also a limitation when denoising images with spatially-varying noise level [61]. Not having a universal denoising model means that multiple models need to be trained and stored for different noise levels, and that noise level knowledge is required at test time. The recent method [60] that generalizes to image restoration tasks is a non-universal non-blind denoiser, where 25 denoising networks are used for noise levels below 50, and even training parameters are chosen based on the noise level. Similarly, Remez *et al.* [39], who reach PSNR results on par with the state of the art, is another non-universal non-blind example. To leverage better priors, images are first classified into a set of classes and every single class has its specific deep model. The method is also not blind and is trained per noise level. Zhang *et al.* [62] present a universal non-blind network for multiple super-resolution degradations by denoising, deblurring, and super-resolving images. They report that although a blind version is more practical, their blind approach fails to perform consistently well since it cannot generalize.

*Blind Universal Denoisers:* The state-of-the-art Gaussian denoiser DnCNN is both universal and blind [59]. It is a deep network that is jointly trained on randomly-sampled noise level patches to generalize denoising to a range of noise levels. It has not been outperformed yet by other methods, whether blind or not [16], [48]. Only the recent FFDNet [61] by the same authors of DnCNN [59] improves on DnCNN for noise levels 50 and 75 by 0.06 and 0.15 $dB$ respectively, on the Berkeley BSD68 set, while performing similarly or worse for other levels. It is, however, not a blind network as it requires a noise level map as input. Lefkimmiatis [26] recently studied universal denoising, building on prior work for modeling patch similarity in CNNs [25]. His methods are, strictly speaking, not universal as two networks are trained separately, one for low ($\leq 30$) and one for high noise levels ($\in [30, 55]$). They are thus non-blind since a noise-level-based choice must be made at inference time. Furthermore, the published results do not outperform the blind DnCNN denoising results. We thus conduct evaluation comparisons of our BUIFD method with the state-of-the-art DnCNN and the classic BM3D approach [8], [9], which is the best non-learning-based denoiser. It leverages image self-similarities by jointly filtering similar image patches. The authors also present a blind version of the BM3D algorithm, and we compare to both blind and non-blind versions.

Our proposed image denoiser BUIFD learns to disentangle its features to predict a prior and a noise level intermediate results. They serve as inputs to the fusion part of the network, responsible for the final denoising. Disentangling the feature space is fundamental for interpretability [6], partial transfer learning [57], domain translation [54], domain adaptation [58], specific attribute manipulation [12], [28], [63] and multi-task networks [2]. In our case, it is fundamental for our theoretical denoising function since the different representations serve as its inputs.

## III. Single-Image Fusion Denoising

### A. Theoretical Framework

Although some specific applications can have a more accurate modeling [24], [49], an additive white Gaussian noise model is often assumed in denoising tasks, as it models common acquisition channels [52]. We thus assume that the additive independent and identically distributed noise $n$ follows a Gaussian distribution $\mathcal{N}(0, \sigma_n^2)$, and is uncorrelated with the data $x$. The noise standard deviation $\sigma_n$ is called noise level. In a Bayesian framework, the conditional probability distribution of the noiseless data $x$ given a noisy observation $y$ (where $y = x + n$) is given by the relation

$$p_{X|Y}(x|y) = \frac{p_{Y,X}(y,x)}{p_Y(y)} = \frac{p_{Y|X}(y|x)p_X(x)}{p_Y(y)}, \qquad (1)$$

where $X$ and $Y$ are the random variables corresponding respectively to $x$ and $y$. We are interested in the conditional distribution as we search for the Maximum Aposteriori Probability (MAP) estimate $\hat{x}$ of $x$. The former is

$$\hat{x} = \arg\max_x p_{X|Y}(x|y). \qquad (2)$$

We also model the data prior on $x$ as a Gaussian distribution $\mathcal{N}(\bar{x}, \sigma_x^2)$ centered at $\bar{x}$ [40]. We later modify this assumption in Sec. III-D to the practical case of real-world images. The conditional probability of $y$ given a noiseless $x$ value is

$$p_{Y|X}(y|x) = \frac{1}{\sqrt{2\pi\,\sigma_n^2}} e^{-\frac{(y-x)^2}{2\sigma_n^2}}, \qquad (3)$$

and the probability distribution of $y$ is the convolution of those of $x$ and $n$, given in the Gaussian case by

$$p_Y(y) = p_X(x) \circledast p_N(n) = \frac{e^{-\frac{(y-\bar{x})^2}{2(\sigma_x^2+\sigma_n^2)}}}{\sqrt{2\pi(\sigma_x^2+\sigma_n^2)}}, \qquad (4)$$

where $\circledast$ is the convolution operator. With these probability distribution functions, we can obtain an expression for the conditional distribution of $x$ given its noisy observation $y$ by substituting Eq. (3) and Eq. (4) into Eq. (1). $p_{X|Y}(x|y)$ can also be written in the following form of a Gaussian in $x$, given an observation $y$

$$p_{X|Y}(x|y) = \frac{1}{\sqrt{2\pi\,\hat{\sigma}_x^2}} e^{-\frac{(x-\hat{\mu})^2}{2\hat{\sigma}_x^2}}. \qquad (5)$$

By matching the expanded expression for $p_{X|Y}(x|y)$ with Eq. (5) for all possible $x$ values, we obtain the expressions for $\hat{\mu}$ and $\hat{\sigma}^2$

$$\hat{\mu} = \frac{\sigma_n^2 \bar{x} + \sigma_x^2 y}{\sigma_x^2 + \sigma_n^2}, \qquad \hat{\sigma}^2 = \frac{\sigma_x^2 \sigma_n^2}{\sigma_x^2 + \sigma_n^2}. \qquad (6)$$

For the Gaussian shown in Eq. (5), the MAP estimator is also the conditional expected value (mode and mean being equal) and it is hence given by

$$\hat{x} = \mathbf{E}[x|y] = \int_{-\infty}^{\infty} x \cdot p_{X|Y}(x|y)dx, \qquad (7)$$

which, using Eq. (5), can be directly derived to be

$$\hat{x} = \frac{\bar{x}}{1+S} + \frac{y}{1+1/S}, \qquad (8)$$

where $S \triangleq \sigma_x^2/\sigma_n^2$ and stands for Signal-to-Noise Ratio (SNR). We call this operation fusion denoising as it fuses the prior and the noisy image, based on the SNR.

Image denoising models are typically trained to maximize PSNR or equivalently minimize Mean Squared Error (MSE)

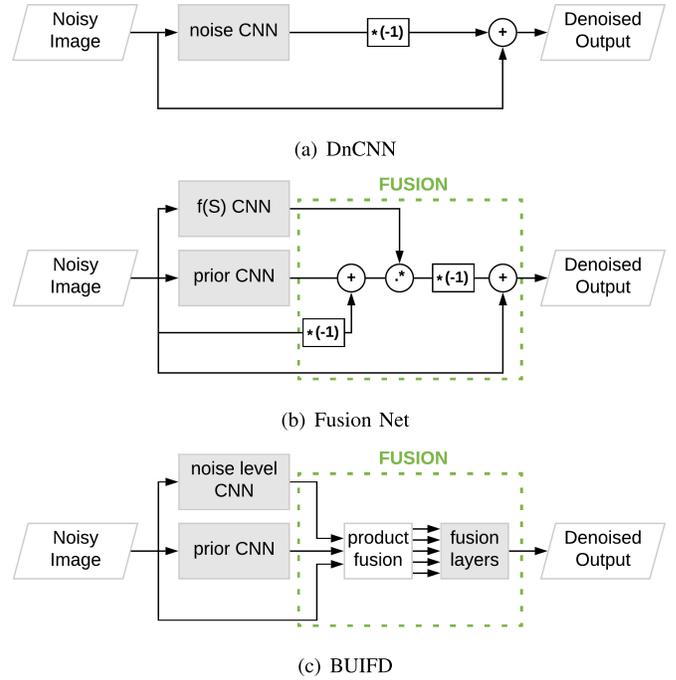

**(a) DnCNN**

**(b) Fusion Net**

**(c) BUIFD**

Fig. 1. (a) Schematic of the DnCNN residual learning approach for denoising. The network predicts the noise in an image. (b) Our Fusion Net jointly learns the SNR function for optimal fusion of the noisy image with the learned prior, following Eq. (8). (c) Our real-image denoiser, BUIFD, where fusion is carried out with a pixel-wise product stage followed by three convolution layers for learning a general fusion function (Sec. III-D).

loss. This means that with close-to-optimal convergence of a neural network model (MSE loss $\to 0^+$), its output tends towards the minimum MSE estimator (MMSE). With our Gaussian modeling, this leads to the MAP estimator $\hat{x}$ of Eq. (8). Thus, an MSE reconstruction loss in a neural network leads it to the estimator $\hat{x}$, iff $S$ and $\bar{x}$ are correctly predicted and correctly used in the fusion with the noisy input $y$, as in Eq. (8). The optimal fusion, used as reference in our experimental evaluation in Sec. IV-B, is given the exact $S$ and $\bar{x}$ values for Eq. (8).

### B. Fusion Net Architecture

We incorporate the basic structure of the optimal fusion solution into the architecture of a neural network, which we call *Fusion Net*. We build the main blocks of our Fusion Net based on the blind DnCNN introduced in [59] and illustrated in Fig. 1(a). In Fig. 1, the noise-predicting CNN of DnCNN (Fig. 1(a)), the prior-predicting CNN, and the one predicting $f(S)$ (where $f(S) \triangleq \frac{1}{1+S}$) in our Fusion Net (Fig. 1(b)), all leverage the same DnCNN architecture design. The CNNs are all constituted of a sequence of convolution layers, rectified linear units (ReLU) [34] and batch normalization blocks [21]. Note that $f(S)$ is inversely-proportional to the SNR and proportional to the noise level. It is the factor multiplying the prior in Eq. (8). To summarize, the $f(S)$ CNN predicts $\frac{1}{1+S}$ where $S$ is the SNR of the input image (determined by the noise level and the image model used in our theoretical settings), and the prior CNN predicts $\bar{x}$ defined in Eq. (7).

Unlike the DnCNN that predicts the *noise* values in the input noisy image, then subtracts them from the noisy input to yield the final denoised output, our network learns optimal *fusion denoising* given by the function in Eq. (8), as illustrated in Fig. 1(b). The same depth and capacity of the DnCNN are retained to learn separately the image prior and the SNR function, $f(S)$, that is required for the weighted fusion of the prior and the noisy input image. Note that SNR learning also contains a form of prior knowledge, but of variance rather than of expectation. We subtract from the prior our noisy input image and multiply the result, pixel-wise, with the SNR function. This yields the noise prediction given a noisy input, which we subtract from the latter to obtain the denoised output. This architecture is mathematically equivalent to Eq. (8). However, the wiring of Fig. 1(b) allows us to clearly have a residual learning connection and to keep the parallelism between the two aforementioned networks.

### C. Fusion Net Feature Disentangling

To mimic the optimal fusion between image prior and noisy image based on the SNR, as in Eq. (8), both the architecture and loss function are adapted. For the fusion, the network needs to predict the image prior $\bar{x}$ and $f(S)$ per pixel (Fig. 1(b)). We obtain, with close-to-zero MSE reconstruction loss of our Fusion Net, that the ground-truth target and the network output are approximately equal

$$\bar{x} \cdot f(S) + y \cdot (1 - f(S)) \approx a \cdot b + y \cdot (1 - b), \quad \forall y \in \mathcal{D}^{\mathcal{T}}, \tag{9}$$

where $a$ and $b$ are the outputs of intermediate layers in the Fusion Net, and $y$ is the noisy input. Specifically, $a$ is the output of the final layer of the prior CNN in Fig. 1(b), and $b$ the output of the last layer of $f(S)$ in the same figure. After gradient descent convergence, when the MSE reconstruction loss is close to zero, we get the approximate equality of the left and right terms in Eq. (9). We can view this equation as a first-degree polynomial in the variable $y$. As Eq. (9) holds for all $y$ in the training dataset $\mathcal{D}^{\mathcal{T}}$, we can apply coefficient equating, where the coefficients are $\{a \cdot b, 1 - b\}$ and $\{\bar{x} \cdot f(S), (1 - f(S))\}$. We thus obtain the approximate equality between $a$ and $\bar{x}$ and between $b$ and $f(S)$. The network intermediate outputs $\{a, b\}$, are therefore, respectively, equal to the prior and the SNR function $\{\bar{x}, f(S)\}$, with close-to-zero MSE reconstruction loss $\forall y \in \mathcal{D}^{\mathcal{T}}$. This extends to other $y$ outside the dataset assuming that the latter is general enough. We can further incorporate optimal denoising information in the Fusion Net, under the theoretical settings described in Sec. III-A, through explicit SNR learning with a dedicated loss term. The fusion representations, i.e. the prior $\bar{x}$ and $f(S)$, are thus further enforced through a penalty term for predicting $f(S)$ in the loss function. The full loss function $\mathcal{L}_f$ of the Fusion Net is given by

$$\mathcal{L}_f = \alpha \|a \cdot b + y \cdot (1 - b) - x\|_2^2 + (1 - \alpha)\|b - f(S)\|_2^2, \tag{10}$$

where $\alpha$ is a weight parameter, the first term is the MSE reconstruction loss similar to that of the DnCNN, and the second term is a reconstruction loss for $f(S)$. Following Eq. (9), $a \cdot b + y \cdot (1 - b)$ is the denoised output of the Fusion Net.

The Fusion Net therefore minimizes the reconstruction loss over the denoised image by learning to predict the image prior and the SNR function values separately. Unlike the DnCNN residual learning network, which only leverages ground-truth noise-free images during training, the Fusion Net also leverages explicit SNR information.

### D. Denoising Non-Gaussian Images

Here, our main objectives are to (1) design a *Blind Universal Image Fusion Denoiser (BUIFD)* for real images, by adapting the theoretical fusion strategy integrated in our Fusion Net, (2) evaluate the denoising performance of BUIFD on training noise levels, and (3) assess the generalization to unseen noise levels with real images.

Since a real image cannot be modeled with a simple Gaussian prior, our image fusion denoising network used for real images (BUIFD), shown in Fig. 1(c), is adapted from the theoretical Fusion Net, shown in Fig. 1(b), by modifying the fusion part. We replace the optimal mathematical fusion by a product fusion step followed by trainable convolution layers. We use three convolution layers to learn the data-dependent fusion function. The optimal fusion function $F$ is to be applied on the noisy input image $y$, the prior prediction, and the noise level prediction

$$\hat{x} = F(y, \ f_P(y, \theta_P), \ f_N(y, \theta_N)), \tag{11}$$

where the prior-predicting and noise-level-predicting network functions are respectively $f_P$ and $f_N$, with their corresponding learned parameters $\theta_P$ and $\theta_N$, and the denoised estimate is $\hat{x}$. Intuitively, the prior-predicting network ($f_P$) is used to predict the expected value of the unknown real-word distribution out of which the intensity of a given pixel is sampled, and that for each pixel. The noise-level-predicting network ($f_N$) predicts the noise level, which is used to control the weighted average between prior and observation. When the noise level is low, the actual observation can be given more weight, and when the noise level is high, the current observation is less reliable and the fusion resorts more to the use of the prior estimation.

The optimal fusion $F$ can be approximated by $\hat{F}$ modeled with three convolution layers. However, we expect $F$ to contain pixel-wise inter-input multiplications similar to the ones of Eq. (8). Since such pixel-wise multiplications cannot be replicated with convolutions, we pass two additional inputs into the convolution layers that model $\hat{F}$. These two additional inputs are given by

$$f_P(y, \theta_P) \odot f_N(y, \theta_N), \quad y \odot (1 - f_N(y, \theta_N)), \tag{12}$$

where $\odot$ is pixel-wise multiplication. They are concatenated with the inputs of $F$ given in Eq. (11), yielding five different inputs that are sent to $\hat{F}$. The two additional inputs reduce the learning burden of the convolution layers and improve the denoising performance. Note that we normalize $f_N(\cdot, \cdot) \in [0, 1]$. We call this pixel-wise multiplication step and the concatenation of the additional inputs the *product fusion* (shown in the pipeline of Fig. 1(c)). These two fusion steps, namely

TABLE I

Test Set PSNR ($dB$) Results for the Noise Standard Deviations Given in the Top Row. The Networks Are Trained on Noise Levels Randomly Chosen in [5, 25]. Noise Levels in the Right Half of the Table Are Not Seen During Training. We Also Report the Optimal Bayesian Denoising (Optimal Fusion). The Bottom Row Shows the Independent Two-Sample t-Test Results Between DnCNN and Our Fusion Net. The Two-Tailed p-Values Validate the Null Hypothesis of Equal Average PSNR Between DnCNN and the Fusion Net on Training Noise Levels, With Significance Level 0.05

| $\sigma$ | 5 | 10 | 15 | 20 | 25 | 30 | 40 | 50 | 60 | 70 |
|---|---|---|---|---|---|---|---|---|---|---|
| Optimal Fusion | 34.325 | 28.778 | 25.947 | 24.261 | 23.185 | 22.464 | 21.604 | 21.138 | 20.860 | 20.681 |
| DnCNN | 34.158 | 28.736 | 25.920 | 24.245 | 23.169 | 22.281 | 20.490 | 18.925 | 17.548 | 16.372 |
| (Ours) Fusion Net | 34.158 | 28.734 | 25.922 | 24.249 | 23.173 | **22.346** | **21.310** | **20.908** | **20.609** | **19.669** |
| p-value | 0.760 | 0.568 | 0.465 | 0.100 | 0.053 | $\approx 0$ | $\approx 0$ | $\approx 0$ | $\approx 0$ | $\approx 0$ |

the product fusion and the three convolution layers, form $\hat{F}$ and realize point (1) above. The BUIFD's optimization loss is given by

$$\mathcal{L}_f = ||\hat{F}(\mathbf{C}) - x||_2^2 + ||f_N(y, \theta_N) - N||_2^2, \quad (13)$$

where $\mathbf{C}$ is the concatenation of the inputs listed in Eq. (11) and Eq. (12), namely, $\{y, f_P(y, \theta_P), f_N(y, \theta_N), f_P(y, \theta_P) \odot f_N(y, \theta_N), y \odot (1 - f_N(y, \theta_N))\}$, $x$ is the ground-truth original image, and $f_N(y, \theta_N)$ and $N$ are respectively the predicted and ground-truth noise level values, normalized to $[0, 1]$. We discuss the relation between BUIFD (Fig. 1(c)) and our theoretical Bayesian network Fusion Net (Fig. 1(b)) in detail in the following section.

### E. Relation With the Bayesian Framework

The Fusion Net in Fig. 1(b) explicitly models the relation with the Bayesian solution in the theoretical experiments. We discuss in what follows the relation between BUIFD (Fig. 1(c)) and the Bayesian solution Eq. (8). We first note that a Gaussian prior does not perfectly model real images, and thus, we expect that the real-image BUIFD network (Fig. 1(c)) deviates from the Fusion Net (Fig. 1(b)), from which it is inspired, to adapt to real images. However, as addressed in Sec. III-D, the relation between BUIFD and the Bayesian framework is strongly pertinent.

First, the product fusion Eq. 12 explicitly creates *the same components* as in the Bayesian equation Eq. (8). This product fusion weighs noisy input and learned prior based on SNR, as in the Bayesian fusion. The fusion layers are only 3 convolutional layers with no non-linearities, to ensure that mostly an additive fusion of our Bayesian terms takes place, with local smoothing, and the relation with the Bayesian solution is preserved as much as possible.

Second, we do not predict an image prior in the sense of a pixel intensity probability distribution, but only the expected mean of that *unknown* distribution. In the literature, priors are often probability distributions of image gradients, but our definition is quite distinctive. *Our prior is, per pixel, the expected value of the distribution out of which the pixel's intensity was sampled.* Even with noise-free images, one cannot exactly know that distribution (nor its mean), per pixel, to assess how much this definition is still respected in the BUIFD network with real images. However, all other Bayesian components are consistent, and the empirical results as well. Our improvement of $3.30dB$ at unseen noise level 70 in the theoretical experiment is paralleled by an improvement of about $3dB$ at noise level 75 in the real image BSD68 experiment.

We hope our methodology motivates future work to analyze deep network optimality on theoretical experiments that are designed such that an optimal solution is known, and that it motivates deep network design inspired from Bayesian solutions.

## IV. Experimental Evaluation

### A. Fusion Net Experimental Setup

The networks are trained (and tested) with data generated synthetically according to the theoretical assumption of a Gaussian image prior as defined in Sec. III-A. The training data is composed of over 200k patches of size $40 \times 40$ pixels. Image pixel intensities for the training data are drawn at random from $\mathcal{N}(127, 25^2)$, following the Gaussian image prior assumption, and all values are normalized to $[0, 1]$ before the training through division by 255 and clipping of all values outside the interval to the interval's closer bound when noise is added. For the testing data, 256 images of size $256 \times 256$ pixels are used, and they are created with the same procedure as that of the training data.

We train the networks for 50 epochs with mini-batches of size 128. We use the Adam optimizer [22] with an initial learning rate of 0.001 that is decayed by a factor of 10 every 30 epochs, the remaining parameters being set to the default values. The weight $\alpha$ in Eq. (10) is set to 0.1. We train the networks with multiple levels of noise. The standard deviation of the additive Gaussian noise is chosen uniformly at random within the interval [5, 25] during the training. At the end of every epoch, the noise components are re-sampled, following the same procedure, but not the ground-truth images. For the testing phase, the networks are evaluated on test images where the added noise is also Gaussian, with a given standard deviation.

### B. Fusion Net Evaluation

PSNR results of DnCNN, our Fusion Net, as well as the optimal upper bound are presented in Table I. The optimal upper bound denoising performance is that of the optimal mathematical solution in Eq. (8). We can see that both the DnCNN and the Fusion Net perform similarly on the training noise levels (left half of the table), and very close to optimal. To validate that the results are indeed statistically similar, we analyze the distribution of PSNR values across the test

TABLE II

PSNR ($dB$)/SSIM COMPARISONS OF *grayscale* IMAGE DENOISING ON THE BSD68 STANDARD TEST SET. WE COMPARE THE NON-BLIND BM3D, THE BLIND BM3D, DnCNN, AND OUR BUIFD. DnCNN$_\sigma$ OR BUIFD$_\sigma$ INDICATES THAT THE NETWORK SEES NOISE LEVELS *only* UP TO $\sigma$ DURING THE TRAINING. BOLD INDICATES THE BEST BLIND RESULT, FOR EACH RANGE OF TESTING NOISE LEVELS, AND THAT BEST RESULT IS SELECTED BEFORE ROUNDING. NOTE: SMALL DEVIATIONS IN REPORTED PSNR VALUES COMPARED WITH THE LITERATURE, NOTABLY ON HIGHER NOISE LEVELS, ARE DUE TO CLIPPING NOISY INPUTS (AND OUTPUTS) TO [0, 255], AS A PRACTICAL CONSIDERATION

| Method | Blind | Test noise level (standard deviation) | | | | | | | |
|---|---|---|---|---|---|---|---|---|---|
| | | 5 | 10 | 15 | 20 | 25 | 30 | 35 | 40 |
| BM3D [8] | No | 37.57/0.964 | 33.27/0.916 | 30.98/0.871 | 29.45/0.831 | 28.32/0.797 | 27.42/0.766 | 26.66/0.739 | 25.98/0.714 |
| | Yes | 29.34/0.806 | 29.18/0.802 | 28.95/0.799 | 28.69/0.798 | 28.32/0.797 | 27.32/0.762 | 25.13/0.638 | 22.39/0.494 |
| DnCNN$_{55}$ [59] | Yes | **37.70/0.967** | **33.61**/0.926 | 31.31/0.882 | 29.65/0.838 | 28.31/0.795 | 27.17/0.754 | 26.19/0.717 | 25.31/0.682 |
| BUIFD$_{55}$ | Yes | 37.49/0.966 | 33.58/**0.926** | **31.40/0.888** | **29.91/0.852** | **28.75/0.819** | **27.80/0.787** | **27.00/0.758** | **26.30/0.731** |
| DnCNN$_{75}$ [59] | Yes | **37.64/0.967** | **33.62/0.927** | **31.37**/0.886 | 29.79/0.844 | 28.55/0.804 | 27.52/0.768 | 26.65/0.736 | 25.84/0.704 |
| BUIFD$_{75}$ | Yes | 37.25/0.964 | 33.47/0.924 | 31.35/**0.886** | **29.88/0.851** | **28.74/0.819** | **27.82/0.788** | **27.01/0.759** | **26.32/0.732** |
| | | 45 | 50 | 55 | 60 | 65 | 70 | 75 | Mean |
| BM3D [8] | No | 25.28/0.686 | 24.79/0.667 | 24.30/0.648 | 23.86/0.632 | 23.43/0.618 | 23.02/0.603 | 22.67/0.591 | 27.13/0.74 |
| | Yes | 20.01/0.389 | 18.22/0.317 | 16.83/0.262 | 15.78/0.222 | 14.86/0.189 | 14.10/0.165 | 13.48/0.147 | 22.17/0.51 |
| DnCNN$_{55}$ [59] | Yes | 24.50/0.648 | 23.75/0.616 | 23.07/0.586 | 22.29/0.546 | 21.06/0.460 | 19.42/0.352 | 17.88/0.278 | 26.08/0.67 |
| BUIFD$_{55}$ | Yes | **25.65/0.704** | **25.06/0.680** | **24.52/0.658** | **23.97/0.637** | **23.31/0.603** | **22.28/0.536** | **20.97/0.451** | **27.20/0.73** |
| DnCNN$_{75}$ [59] | Yes | 25.14/0.675 | 24.48/0.647 | 23.90/0.621 | 23.34/0.597 | 22.87/0.577 | 22.41/0.558 | 22.01/0.541 | 27.01/0.72 |
| BUIFD$_{75}$ | Yes | **25.68/0.706** | **25.11/0.682** | **24.55/0.658** | **24.03/0.636** | **23.56/0.617** | **23.10/0.598** | **22.66/0.582** | **27.37/0.75** |

set. A two-sided T-test (independent two-sample T-test) is used to evaluate the null hypothesis that the PSNR results of both networks have similar expected values. This test is chosen as we have the exact same sample sizes defined by the test dataset, and the variances of PSNR results are very similar. The T-test results are given in the bottom row of Table I, and the null hypothesis holds for all configurations in the left half of the table (for a 0.05 significance level, i.e., a $p$-value $\geq 0.05$). This shows that the Fusion Net, despite the modeling that mimics optimal denoising fusion and the additional training information to learn SNR values, performs similarly to the DnCNN. The latter has therefore enough capacity and learns an optimal denoising. This, however, only holds for the noise levels seen during training by the networks, shown in the left half of Table I. The confidence in the null hypothesis decreases with increasing test noise levels. With a significance level above 0.053, the null hypothesis would even be rejected for noise level 25.

The evaluation results on noise levels larger than 25, which are not trained on by any of the networks, are reported in the right half of Table I. For these larger noise levels, the null hypothesis is very clearly rejected as there is a growing performance gap between DnCNN and our Fusion Net. The $p$-value quickly drops to zero when there is a PSNR gap, since variances are very small in our results. The Fusion Net generalizes better to unseen noise levels, even performing close to optimal up to noise level 60. The further we increase the noise level, the larger is the performance gap between the Fusion Net and the DnCNN. Although both networks perform well for the training noise levels, the Fusion Net learns a more general model and clearly outperforms on unseen noise levels.

### C. Real-Image Experimental Setup

We use the referenced implementation by the authors of DnCNN and the same datasets.[1] As mentioned in Sec. III-D, the architecture of our prior-predicting network is identical to

TABLE III

WE EVALUATE PSNR VALUES, WITH SPATIALLY-VARYING NOISE LEVEL, ON THE BSD68 TEST SET. THE NOISE LEVEL INCREASES LINEARLY WITHIN THE IMAGE OVER THE RANGE $[\sigma_c - 10, \sigma_c + 10]$. THE NON-BLIND BM3D IS GIVEN THE CENTRAL NOISE LEVEL $\sigma_c$

| $\sigma_c$ | 15 | 25 | 40 | 55 | 65 |
|---|---|---|---|---|---|
| Non-blind BM3D | 29.30 | 27.80 | 25.75 | 24.28 | 23.41 |
| BM3D | 28.94 | 27.80 | 21.63 | 16.78 | 14.85 |
| DnCNN$_{55}$ | 31.24 | 28.32 | 25.41 | 23.17 | 20.83 |
| **BUIFD$_{55}$** | **31.38** | **28.74** | **26.22** | **24.33** | **22.81** |
| DnCNN$_{75}$ | 31.31 | 28.51 | 25.80 | 23.87 | 22.83 |
| **BUIFD$_{75}$** | **31.34** | **28.73** | **26.29** | **24.52** | **23.53** |

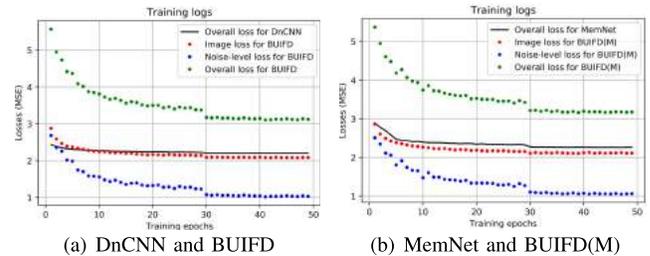

(a) DnCNN and BUIFD     (b) MemNet and BUIFD(M)

Fig. 2. Training losses of the different learning-based methods. Per epoch, we plot with a full black curve the overall loss (i.e. reconstruction loss) of the base methods DnCNN and MemNet, in (a) and (b) respectively. The same reconstruction loss with our fusion method is plotted with a dotted red curve, the noise-level loss computed on the corresponding intermediate output (i.e. the output of the noise level CNN) is plotted with a dotted blue curve, and the overall loss for the fusion methods (the sum of the former two losses) is plotted with a dotted green curve. Note the abrupt small improvement in loss reduction at epoch 30, which is when the learning rate is exponentially decayed. We can see that the different learned function converge by the end of training (logs shown for the methods with upper training noise level 55).

that of DnCNN.[2] All the network details are available in [59] and we omit the repetition. The same network depth and feature layers are thus used in the prior-predicting network (18 main blocks) in Fig. 1(c). The noise level network is a shallower one consisting of 5 blocks similar to the ones

---

[1] https://github.com/SaoYan/DnCNN-PyTorch

[2] Our code is available at: https://github.com/majedelhelou/BUIFD

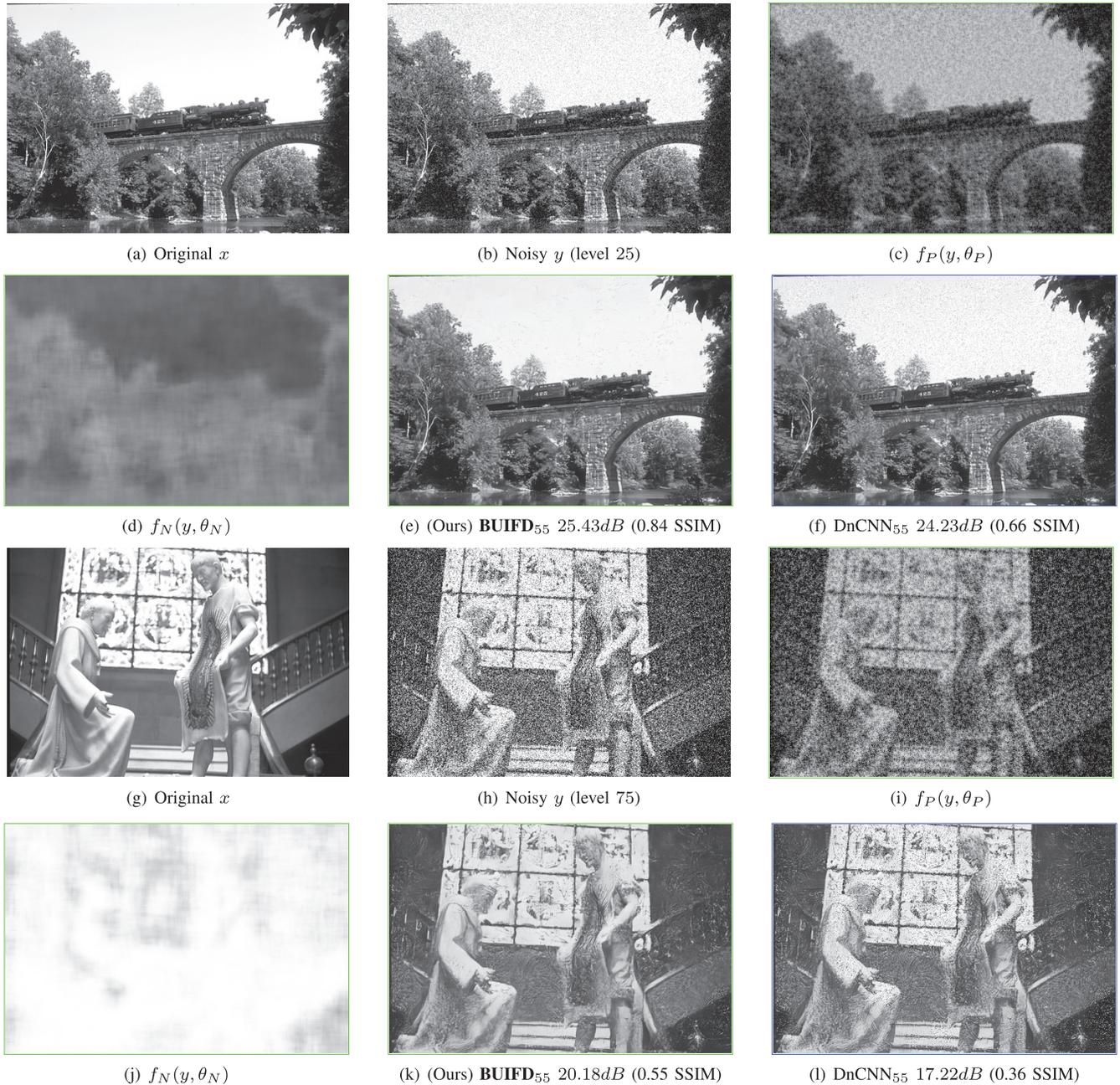

Fig. 3. Left to right: original and noisy images, prior and noise level predictions of BUIFD, our fused denoising result and the DnCNN denoised image. Our denoising result is created by fusing the noisy image, the prior and the noise level values, for instance (e) is $\hat{F}((b), (c), (d))$. All the networks are trained on noise levels in $[0, 55]$. Whether the noise level is seen (25), or not seen (75), during training, our denoised results show better noise removal (sky in (e-f), window, wall and arms in (k-l)). We show the PSNR in $dB$ and the SSIM [50] between parentheses for the different results. Best viewed on screen.

used in the prior predictor. Each block is a convolution followed by a batch normalization and a ReLU, and we append to the noise level predictor a convolution followed by an application of the logistic sigmoid function to obtain the normalized $f_N(\cdot, \cdot) \in [0, 1]$. The noise level values are thus mapped during the training to the range $[0, 1]$ by dividing by the largest *training* noise level. The three convolution layers approximating the final fusion have 16 channels. Both the BUIFD and the DnCNN networks are trained with the same training parameters and optimization settings, similar to Sec. IV-A except for the patch size. For completeness, we provide all the details of the training hyper-parameters. We use the Adam optimizer [22] with an initial learning rate of 0.001 that is decayed by a factor of 10 every 30 epochs, the remaining optimizer parameters being set to the default values. The networks are trained for 50 epochs each, and the progress of the different losses can be seen in Fig. 2. We use a patch size of $50 \times 50$ with a stride of 10 on the training images. The training mini-batch size is set to 128 patches per mini-batch. The added noise is drawn from a Gaussian distribution of given standard deviation, based on the noise level. This standard deviation is sampled uniformly at random

TABLE IV

PSNR ($dB$)/SSIM COMPARISONS OF *Color* IMAGE DENOISING, SIMILAR TO TABLE II, ON THE CBSD68 STANDARD TEST SET. BOLD INDICATES THE BEST BLIND RESULT, FOR EACH RANGE OF TRAINING NOISE LEVELS, AND THAT BEST RESULT IS SELECTED BEFORE ROUNDING

| Method | Blind | Test noise level (standard deviation) | | | | | | | |
|---|---|---|---|---|---|---|---|---|---|
| | | 5 | 10 | 15 | 20 | 25 | 30 | 35 | 40 |
| CBM3D [9] | No | 40.19/0.979 | 35.75/0.950 | 33.26/0.919 | 31.52/0.888 | 30.18/0.859 | 29.07/0.830 | 28.09/0.801 | 27.18/0.771 |
| | Yes | 28.17/0.772 | 28.08/0.769 | 27.94/0.765 | 27.74/0.760 | 27.49/0.754 | 27.21/0.748 | 26.90/0.743 | 26.58/0.738 |
| CDnCNN$_{55}$ [59] | Yes | 40.05/0.979 | 35.92/0.953 | 33.57/0.927 | 31.93/0.902 | 30.66/0.877 | 29.61/0.853 | 28.71/0.830 | 27.92/0.808 |
| CBUIFD$_{55}$ | Yes | **40.07/0.979** | **36.01/0.955** | **33.66/0.930** | **32.02/0.905** | **30.75/0.881** | **29.72/0.858** | **28.81/0.835** | **28.01/0.813** |
| CDnCNN$_{75}$ [59] | Yes | 39.75/0.978 | 35.74/0.953 | 33.46/0.928 | 31.86/0.903 | 30.61/0.879 | 29.59/0.855 | 28.70/0.833 | 27.92/0.812 |
| CBUIFD$_{75}$ | Yes | **40.05/0.980** | **35.98/0.955** | **33.65/0.930** | **32.03/0.906** | **30.76/0.883** | **29.71/0.860** | **28.81/0.838** | **28.01/0.816** |
| | | 45 | 50 | 55 | 60 | 65 | 70 | 75 | Mean |
| CBM3D [9] | No | 26.53/0.751 | 25.85/0.729 | 25.21/0.708 | 24.62/0.689 | 24.05/0.670 | 23.51/0.653 | 22.99/0.637 | 28.53/0.79 |
| | Yes | 26.23/0.733 | 25.85/0.729 | 25.41/0.720 | 24.83/0.695 | 24.05/0.647 | 23.07/0.581 | 21.93/0.508 | 26.10/0.71 |
| CDnCNN$_{55}$ [59] | Yes | 27.16/0.786 | 26.49/0.766 | 25.84/0.747 | 25.23/0.729 | 24.65/0.713 | 24.09/0.697 | 23.52/0.677 | 29.02/0.82 |
| CBUIFD$_{55}$ | Yes | **27.27/0.793** | **26.59/0.773** | **25.94/0.754** | **25.33/0.737** | **24.75/0.720** | **24.18/0.703** | **23.62/0.684** | **29.11/0.82** |
| CDnCNN$_{75}$ [59] | Yes | 27.19/0.792 | 26.52/0.772 | 25.89/0.753 | 25.27/0.735 | 24.69/0.717 | 24.13/0.701 | 23.59/0.684 | 28.99/0.82 |
| CBUIFD$_{75}$ | Yes | **27.28/0.796** | **26.60/0.776** | **25.96/0.758** | **25.34/0.740** | **24.76/0.722** | **24.18/0.705** | **23.64/0.689** | **29.12/0.82** |

from a specified range (details in Sec. IV-D), and is the same for all pixels in a given training patch. We use the training hyper-parameters of DnCNN, for training it and for training BUIFD, the hyper-parameters are not tweaked for BUIFD. The noise level predictor is jointly trained within BUIFD, so both network branches always see the same training data (with the same simulated noise distributions) as each other in the experiments of Sec IV-D. We use the 400 Berkeley images [7], [43] for grayscale training and the 432 color Berkeley images for color training, as in [59]. The same architectures are retained for grayscale and color networks.

### D. Real-Image Evaluation

Grayscale denoising evaluation is carried out over the standard Berkeley 68 image test set (BSD68) [41] taken from [32]. Table II reports the results of our fusion approach and of the state-of-the-art blind DnCNN, when they are both trained with noise levels up to 55 or up to 75. Note that for our fusion approach that is trained up to noise level 55, we map the maximum network prediction of 1, during training, to 55 and not to the maximum test noise level, for a more fair comparison. The results of the blind version of BM3D as well as those of the non-blind BM3D, which is given the correct test noise level at inference time, are also reported for reference. We restrict all noisy test images to the range [0, 255], as having negative intensities, or values exceeding 255, is not a configuration encountered in practice.

Fig. 3 shows our intermediate feature results, the prior and the noise level values, along with denoising results. The denoised image is created by fusing the noisy input image with the network-derived prior and the noise level values. The fusion is carried out by the product fusion step and the three convolution layers. As in practical scenarios, the denoised outputs are clipped to [0, 255], as are the noisy input images. Our results better remove the noise compared with those of DnCNN over low frequency regions, and details are better reconstructed over the high-frequency content. We note that, at high noise levels, there is a smudging effect most visible around low-frequency regions (Fig. 3 (k) and (l)), which creates blurry and noisy edges. These are created by both networks, but are more salient in our result (k) as it is less noisy than (l). The higher the noise level and standard deviation of the Gaussian noise, the larger the number of averaged samples needs to be such that the statistical mean converges to zero. This makes the local mean of the noise across small patches vary around zero from region to region, randomly, and causes the smudging-like or wave-like effect (notice over low-frequency regions how almost all these artifacts have a curve shape, rather than a linear one, which is modeled by the various different mean values around them).

As seen in Table II, our fusion approach improves the PSNR at every single noise level starting from $15-20$, which includes seen levels for both training ranges. Comparing DnCNN$_{75}$ and BUIFD$_{75}$, which are trained on all noise levels, we also note with our approach an improvement of up to $0.7\,dB$ and an average improvement of $0.36\,dB$. We outperform even the non-blind version of BM3D by an average of $0.25\,dB$ with our version trained on all noise levels and we perform just as well as the non-blind BM3D when training only up to level 55. Comparing the results of DnCNN$_{55}$ and BUIFD$_{55}$ in Table II, for unseen noise levels in the range (55, 75], we see that the generalization of the fusion approach to unseen noise levels indeed applies to real images. The improvement of multiple PSNR points for level 75 is consistent with that obtained in our synthetic experiment in Table I.

The results in Table III illustrate denoising images with spatially-varying noise level, without re-training the networks. Noise is added across an image with a level that increases linearly with rows. For the non-blind BM3D, we input the average noise level as a guide. The BUIFD network can handle spatially-varying noise, which neither the prior nor the noise level predicting network branches are trained on. It outperforms DnCNN on all noise setups, whether the networks are trained on the full range or only up to level 55.

For color image denoising, we use the standard color version of BSD68 (CBSD68) for testing. Noise is simulated and added to each test image before running it through a denoising method. PSNR results are reported in Table IV. The high inter-channel correlation between the RGB color channels [13] allows all methods to perform significantly better in terms of denoising PSNR on color images compared

TABLE V
PSNR/SSIM EVALUATION OF THE *Blind* BM3D, EPLL, KSVD, WNNM, DnCNN, BUIFD, MemNet, AND BUIFD(M). BOLD INDICATES THE BEST BLIND DENOISING RESULT IN TERMS OF PSNR OR SSIM BETWEEN EACH PAIR OF LEARNING METHODS, FOR EACH GAUSSIAN NOISE LEVEL. WE CLIP NOISY IMAGES TO [0, 255], AS A PRACTICAL CONSIDERATION

| Dataset | Method | 10 | 20 | 30 | 40 | 50 | 60 | 70 | 80 |
|---|---|---|---|---|---|---|---|---|---|
| | | | | | PSNR/SSIM average results per test noise level | | | | |
| BSD68 | KSVD [1] | 27.10/0.713 | 27.71/0.750 | 26.56/0.715 | 21.48/0.444 | 18.12/0.297 | 15.85/0.212 | 14.28/0.165 | 13.09/0.134 |
| | BM3D [8] | 29.18/0.802 | 28.69/0.798 | 27.35/0.763 | 22.44/0.495 | 18.19/0.316 | 15.73/0.220 | 14.10/0.166 | 12.90/0.132 |
| | EPLL [64] | 29.51/0.798 | 29.14/0.808 | 26.07/0.707 | 20.82/0.430 | 17.54/0.289 | 15.70/0.215 | 14.53/0.175 | 13.61/0.149 |
| | WNNM [17] | 27.83/0.750 | 28.35/0.779 | 27.06/0.743 | 21.89/0.468 | 18.22/0.306 | 15.83/0.217 | 14.21/0.166 | 13.00/0.134 |
| | DnCNN [59] | **33.61/0.926** | 29.65/0.838 | 27.17/0.754 | 25.31/0.682 | 23.75/0.616 | 22.29/0.546 | 19.42/0.352 | 16.67/0.233 |
| | BUIFD | 33.58/**0.926** | **29.91/0.852** | **27.80/0.787** | **26.30/0.731** | **25.06/0.680** | **23.97/0.637** | **22.28/0.536** | **19.63/0.374** |
| | MemNet [47] | 33.33/0.927 | 29.59/0.848 | 27.32/0.769 | 25.63/0.701 | 24.35/0.646 | 23.34/0.606 | 21.53/0.499 | 18.43/0.320 |
| | BUIFD(M) | **33.59/0.928** | **29.90/0.856** | **27.83/0.794** | **26.36/0.740** | **25.15/0.690** | **24.14/0.655** | **22.14/0.537** | **18.92/0.363** |
| Set5 | KSVD [1] | 29.03/0.804 | 29.41/0.807 | 27.71/0.738 | 22.10/0.452 | 18.58/0.292 | 16.29/0.207 | 14.64/0.158 | 13.33/0.127 |
| | BM3D [8] | 32.19/0.882 | 30.98/0.856 | 28.72/0.791 | 23.12/0.509 | 18.73/0.324 | 16.24/0.222 | 14.50/0.160 | 13.15/0.125 |
| | EPLL [64] | 32.11/0.865 | 31.04/0.849 | 27.22/0.728 | 21.39/0.434 | 17.93/0.279 | 16.04/0.204 | 14.77/0.164 | 13.75/0.138 |
| | WNNM [17] | 30.16/0.840 | 30.50/0.839 | 28.39/0.765 | 22.51/0.476 | 18.69/0.305 | 16.29/0.216 | 14.59/0.161 | 13.25/0.127 |
| | DnCNN [59] | **35.09/0.927** | 30.83/0.839 | 27.84/0.758 | 25.38/0.683 | 23.47/0.614 | 21.85/0.545 | 19.48/0.372 | 16.88/0.250 |
| | BUIFD | 35.08/**0.928** | **31.56/0.871** | **29.18/0.815** | **27.28/0.765** | **25.68/0.719** | **24.17/0.672** | **22.44/0.580** | **20.16/0.431** |
| | MemNet [47] | 34.67/0.921 | 30.95/0.857 | 28.00/0.773 | 25.79/0.707 | 23.99/0.652 | 22.62/0.614 | 21.20/0.531 | 18.71/0.365 |
| | BUIFD(M) | **35.07/0.929** | **31.56/0.873** | **29.25/0.821** | **27.48/0.775** | **25.94/0.733** | **24.51/0.697** | **22.69/0.606** | **19.80/0.445** |
| Set14 | KSVD [1] | 27.73/0.729 | 28.19/0.755 | 26.74/0.711 | 21.50/0.447 | 18.12/0.304 | 15.90/0.219 | 14.28/0.169 | 13.08/0.138 |
| | BM3D [8] | 30.58/0.832 | 29.68/0.818 | 27.85/0.772 | 22.55/0.502 | 18.24/0.326 | 15.82/0.231 | 14.14/0.172 | 12.91/0.137 |
| | EPLL [64] | 30.45/0.814 | 29.70/0.815 | 26.28/0.707 | 20.89/0.435 | 17.58/0.298 | 15.76/0.221 | 14.54/0.179 | 13.61/0.152 |
| | WNNM [17] | 28.89/0.777 | 29.24/0.796 | 27.49/0.748 | 22.00/0.476 | 18.28/0.316 | 15.92/0.227 | 14.24/0.173 | 13.01/0.138 |
| | DnCNN [59] | **33.81/0.914** | 29.98/0.832 | 27.39/0.757 | 25.40/0.688 | 23.66/0.625 | 22.09/0.553 | 19.35/0.364 | 16.54/0.239 |
| | BUIFD | 33.73/**0.914** | **30.34/0.852** | **28.18/0.795** | **26.55/0.742** | **25.23/0.694** | **23.98/0.650** | **22.33/0.556** | **19.69/0.391** |
| | MemNet [47] | 33.45/0.912 | 29.91/0.842 | 27.43/0.767 | 25.56/0.701 | 24.17/0.649 | 23.06/0.609 | 21.34/0.506 | 18.29/0.331 |
| | BUIFD(M) | **33.70/0.914** | **30.29/0.854** | **28.19/0.801** | **26.62/0.751** | **25.33/0.706** | **24.19/0.670** | **22.27/0.557** | **19.08/0.388** |
| Sun_Hays80 | KSVD [1] | 28.80/0.767 | 29.22/0.778 | 27.21/0.681 | 21.45/0.374 | 18.05/0.233 | 15.82/0.161 | 14.24/0.122 | 13.03/0.097 |
| | BM3D [8] | 31.35/0.848 | 30.63/0.837 | 28.93/0.787 | 23.11/0.465 | 18.44/0.268 | 15.89/0.175 | 14.20/0.126 | 12.94/0.097 |
| | EPLL [64] | 31.09/0.826 | 30.76/0.833 | 27.16/0.710 | 21.12/0.380 | 17.68/0.236 | 15.79/0.167 | 14.57/0.133 | 13.63/0.110 |
| | WNNM [17] | 29.89/0.795 | 30.30/0.810 | 28.56/0.750 | 22.43/0.416 | 18.51/0.246 | 16.06/0.164 | 14.36/0.120 | 13.09/0.094 |
| | DnCNN [59] | 34.94/0.933 | 31.08/0.853 | 28.48/0.771 | 26.24/0.689 | 24.33/0.617 | 22.55/0.535 | 19.53/0.306 | 16.63/0.183 |
| | BUIFD | **34.99/0.935** | **31.44/0.871** | **29.36/0.814** | **27.77/0.763** | **26.41/0.716** | **25.14/0.674** | **23.24/0.561** | **20.17/0.364** |
| | MemNet [47] | 34.65/0.932 | 31.07/0.864 | 28.74/0.792 | 26.88/0.726 | 25.39/0.670 | 24.17/0.629 | 22.34/0.520 | 18.78/0.297 |
| | BUIFD(M) | **34.97/0.935** | **31.42/0.872** | **29.39/0.819** | **27.86/0.771** | **26.55/0.728** | **25.37/0.696** | **23.19/0.575** | **19.44/0.373** |
| Urban100 | KSVD [1] | 27.49/0.793 | 27.93/0.808 | 26.05/0.726 | 21.21/0.487 | 18.05/0.353 | 15.88/0.274 | 14.32/0.224 | 13.12/0.188 |
| | BM3D [8] | 30.98/0.884 | 29.93/0.868 | 27.87/0.818 | 22.64/0.565 | 18.43/0.387 | 15.77/0.287 | 14.30/0.227 | 13.04/0.189 |
| | EPLL [64] | 30.06/0.857 | 29.16/0.851 | 25.99/0.748 | 20.90/0.489 | 17.73/0.354 | 15.89/0.280 | 14.65/0.236 | 13.69/0.205 |
| | WNNM [17] | 28.03/0.796 | 28.54/0.805 | 27.16/0.768 | 21.93/0.529 | 18.33/0.367 | 15.98/0.266 | 14.33/0.208 | 13.08/0.168 |
| | DnCNN [59] | **34.10/0.935** | 30.01/0.870 | 27.10/0.797 | 24.76/0.723 | 22.87/0.656 | 21.17/0.579 | 18.84/0.414 | 16.41/0.303 |
| | BUIFD | 33.72/0.933 | **30.18/0.882** | **27.86/0.833** | **26.04/0.783** | **24.54/0.736** | **23.69/0.698** | **21.67/0.597** | **19.45/0.454** |
| | MemNet [47] | 33.46/0.930 | 29.65/0.869 | 26.89/0.799 | 24.81/0.734 | 23.25/0.679 | 22.10/0.638 | 20.61/0.547 | 18.17/0.396 |
| | BUIFD(M) | **33.63/0.933** | **30.03/0.881** | **27.73/0.832** | **25.99/0.785** | **24.57/0.742** | **23.36/0.704** | **21.64/0.602** | **19.00/0.455** |
| Manga109 | KSVD [1] | 29.91/0.871 | 29.69/0.868 | 27.08/0.763 | 22.02/0.519 | 18.70/0.355 | 16.42/0.245 | 14.73/0.187 | 13.45/0.152 |
| | BM3D [8] | 33.45/0.924 | 31.52/0.910 | 28.80/0.858 | 23.54/0.607 | 19.20/0.428 | 16.63/0.296 | 14.79/0.208 | 13.39/0.160 |
| | EPLL [64] | 33.31/0.915 | 31.29/0.905 | 27.22/0.795 | 21.73/0.531 | 18.34/0.365 | 16.31/0.252 | 14.93/0.198 | 13.87/0.167 |
| | WNNM [17] | 31.58/0.870 | 31.31/0.872 | 28.80/0.803 | 22.83/0.511 | 18.93/0.335 | 16.48/0.234 | 14.73/0.171 | 13.41/0.132 |
| | DnCNN [59] | 35.57/0.936 | 30.50/0.831 | 26.78/0.725 | 23.98/0.638 | 21.82/0.569 | 20.03/0.493 | 17.88/0.331 | 15.74/0.236 |
| | BUIFD | **35.88/0.947** | **31.86/0.907** | **29.09/0.864** | **26.92/0.820** | **25.13/0.777** | **23.58/0.731** | **21.89/0.632** | **19.63/0.480** |
| | MemNet [47] | 34.88/0.940 | 30.58/0.867 | 27.34/0.777 | 24.92/0.699 | 23.11/0.641 | 21.79/0.603 | 20.28/0.512 | 18.05/0.369 |
| | BUIFD(M) | **35.81/0.948** | **31.84/0.912** | **29.19/0.878** | **27.15/0.848** | **25.45/0.822** | **23.97/0.795** | **22.13/0.692** | **19.53/0.556** |

with grayscale images. We note that this advantage of having multiple correlated channels as in color imaging is not always available, for instance with single-wavelength imaging [29]. We hypothesize that this correlation also enables the networks to implicitly learn the noise level prediction. High correlation implies that the network sees multiple approximately equal data samples with different noise instances drawn from the same distribution. Thus, it more easily learns an estimate of the noise variance. Each of the two networks therefore performs more or less the same when trained up to noise level 55 and when trained up to noise levels 75. Our fusion approach, however, consistently outperforms CDnCNN on every single noise level for both training noise ranges. Our average improvement over CDnCNN is about $0.1 dB$. We also note that the networks outperform, on average, even the non-blind CBM3D by about $0.5 dB$ for CDnCNN and $0.6 dB$ for our CBUIFD.

Sample image denoising results for grayscale and color images are illustrated in Fig. 4, 5 and Fig. 6, 7 respectively, for the non-blind BM3D and the blind networks DnCNN and BUIFD trained on the full range of noise levels. The main trade-off seen between the results of BM3D and those of DnCNN is on detail reconstruction. The non-blind BM3D achieves good PSNR reconstruction but at the expense of blurring the results. This causes a loss of details (visible on the large rock in Fig. 4, and the zoom-in insert in Fig. 5)

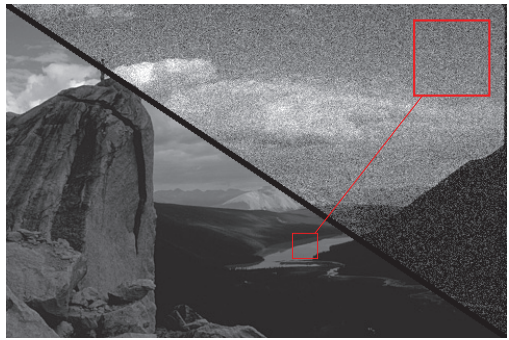

(a) Original\Noisy (level 25)

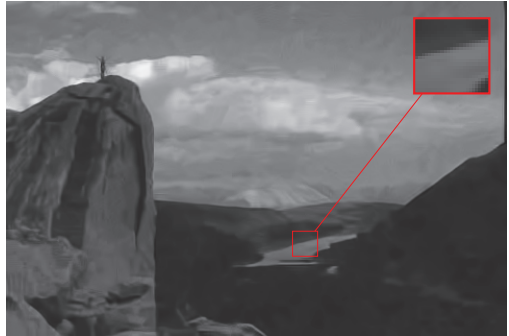

(b) BM3D (non-blind) $31.71dB$ (0.834 SSIM)

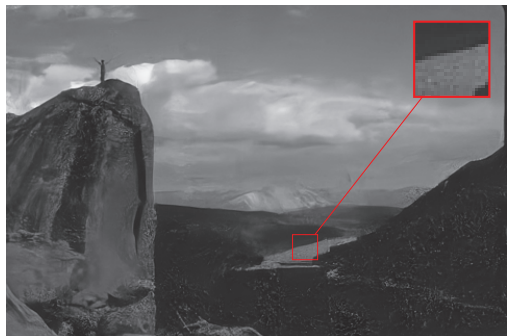

(c) DnCNN$_{75}$ $31.25dB$ (0.792 SSIM)

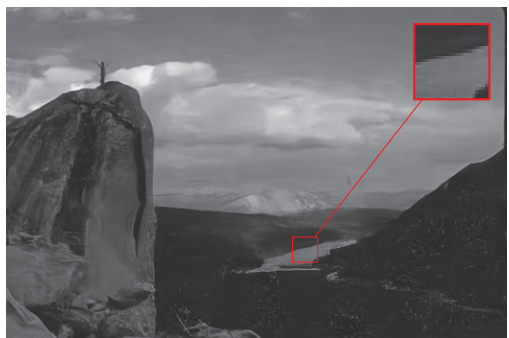

(d) (Ours) **BUIFD**$_{75}$ $31.81dB$ (0.841 SSIM)

Fig. 4. Grayscale image denoising example from BSD68. All networks are trained on all noise levels [0, 75] and we test on noise level 25. Non-blind BM3D loses edge details due to blur smoothing. The network results are sharper, with the better PSNR being that of BUIFD$_{75}$. Best viewed on screen.

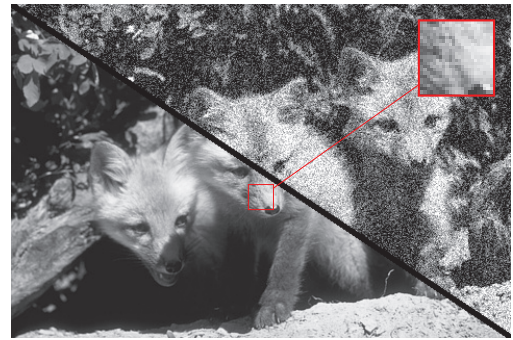

(a) Original\Noisy (level 45)

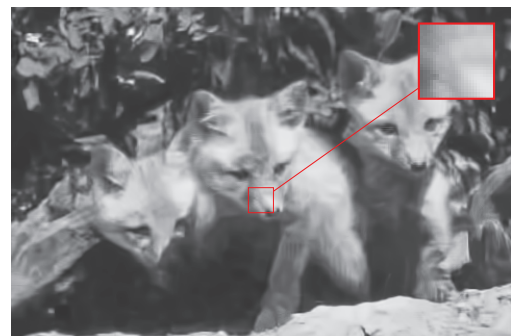

(b) BM3D (non-blind) $24.31dB$ (0.675 SSIM)

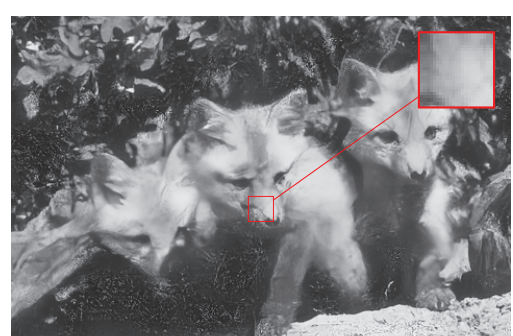

(c) DnCNN$_{75}$ $23.67dB$ (0.618 SSIM)

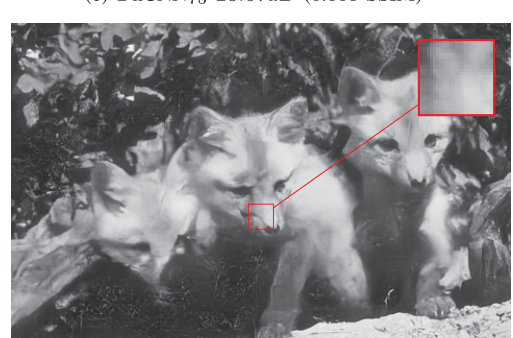

(d) (Ours) **BUIFD**$_{75}$ $24.43dB$ (0.677 SSIM)

Fig. 5. Grayscale image denoising example from BSD68. All networks are trained on all noise levels [0, 75] and we test on noise level 45. Non-blind BM3D results are very smoothed, and details are lost. DnCNN preserves more details, but at the expense of PSNR. Our blind approach preserves details and outperforms the non-blind BM3D in terms of PSNR. Best viewed on screen.

and of edge sharpness (visible on the borders of the lake in the zoom-in insert in Fig. 4). The DnCNN results suffer less of a blurring problem, but the noise-removal is not optimal in certain areas such as smooth surfaces (visible on the inner area of the lake in the zoom-in insert in Fig. 4). Our approach achieves a good performance in terms of this trade-off. BUIFD achieves good PSNR results, with significantly less blurring than the non-blind BM3D (see Fig. 5 for example).

### E. Extended Benchmark Comparisons

We present more denoising experimental tests on different benchmark datasets, and compare the results of different

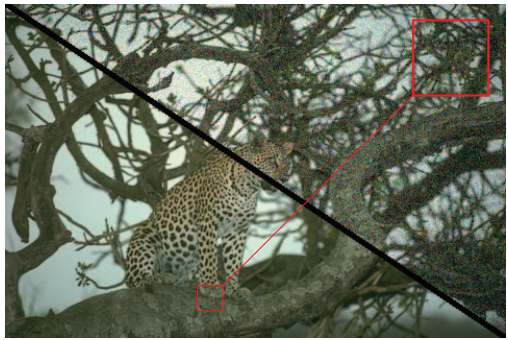

(a) Original\Noisy (level 25)

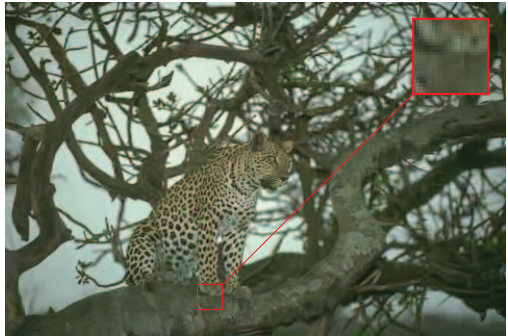

(b) CBM3D (non-blind) $29.81dB$ (0.852 SSIM)

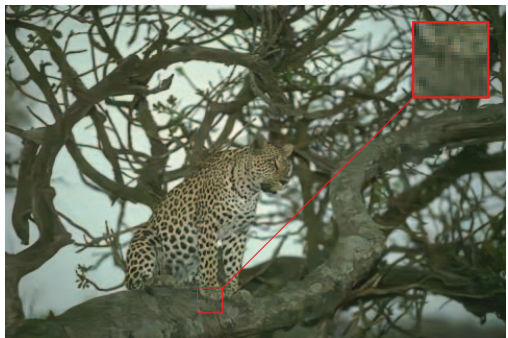

(c) CDnCNN$_{75}$ $30.44dB$ (0.878 SSIM)

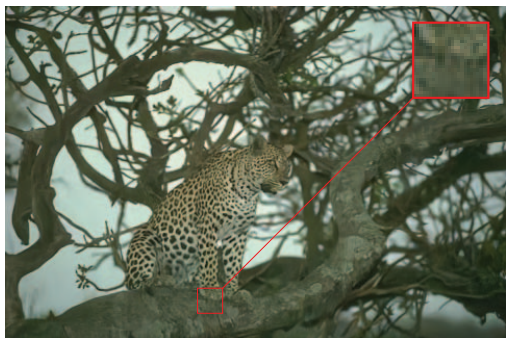

(d) (Ours) **CBUIFD**$_{75}$ $30.62dB$ (0.880 SSIM)

Fig. 6. Color image denoising example from CBSD68. All networks are trained on the full range of noise levels $[0, 75]$ and we test on noise level 25. Best viewed on screen.

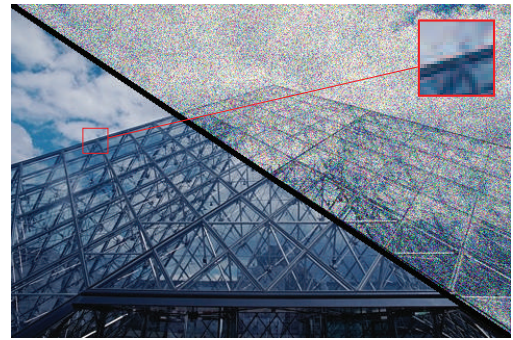

(a) Original\Noisy (level 45)

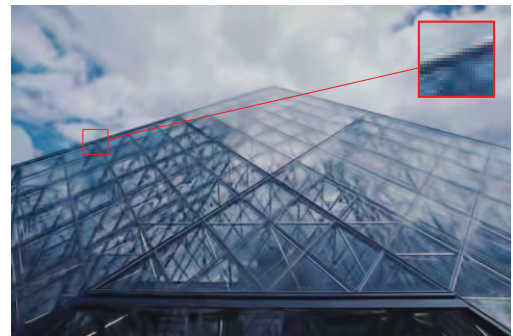

(b) CBM3D (non-blind) $25.79dB$ (0.711 SSIM)

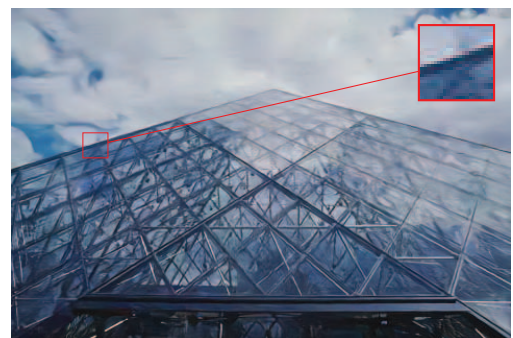

(c) CDnCNN$_{75}$ $26.43dB$ (0.764 SSIM)

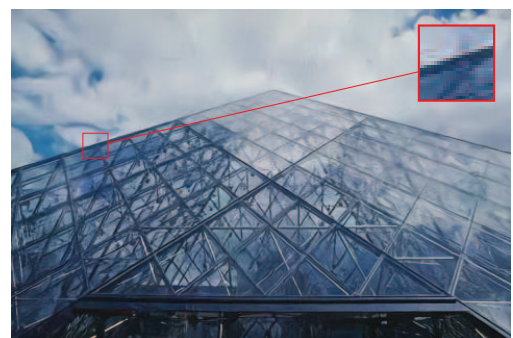

(d) (Ours) **CBUIFD**$_{75}$ $26.68dB$ (0.775 SSIM)

Fig. 7. Color image denoising example from CBSD68. All networks are trained on the full range of noise levels $[0, 75]$ and we test on noise level 45. Best viewed on screen.

denoising approaches on these datasets. We report *blind* denoising results for noise levels 10 to 80 (with a step size of 10) on the BSD68 dataset, Set5, Set14, Sun_Hays80, Urban100, and Manga109 datasets. Set5 and Set14 are made up of, respectively, 5 and 14 traditionally-used images for testing image processing algorithms. Most of their images are smaller than $512 \times 512$. The Sun_Hays80 dataset is made up of the high-resolution version of the 80 images presented in [46], with sizes smaller than $1024 \times 1024$. The Urban100 dataset is a collection of 100 high resolution images taken from Flickr using urban keywords [20]. The Manga109 dataset is constituted of 109 professional artist drawings [33], of size $827 \times 1170$. We present in Table V the denoising results of the blind non-learning methods BM3D, EPLL [64], KSVD [1], and WNNM [17] that were developed for Gaussian denoising and are given, to enforce the blind setting, the default noise level set by the non-blind BM3D (set to 25), and the learning-based methods DnCNN [59] and BUIFD, on denoising the luminance of the images with added Gaussian noise levels ranging from 10 to 80. We also evaluate another learning-based method with the same training hyper-parameters as those of DnCNN, namely, the MemNet architecture [47], and extend our fusion technique to that architecture and call it BUIFD(M). It is constructed following Fig. 1(c), with the exception that the MemNet architecture replaces that of DnCNN for the prior-predicting CNN. All the learning-based methods in this section are trained up to noise level 55. Table V shows the PSNR and SSIM metrics for each method, and we highlight in bold the best-PSNR and best-SSIM method between DnCNN and BUIFD, and between MemNet and BUIFD(M). A sample visual result is shown in Fig. 8, taken from Set14.

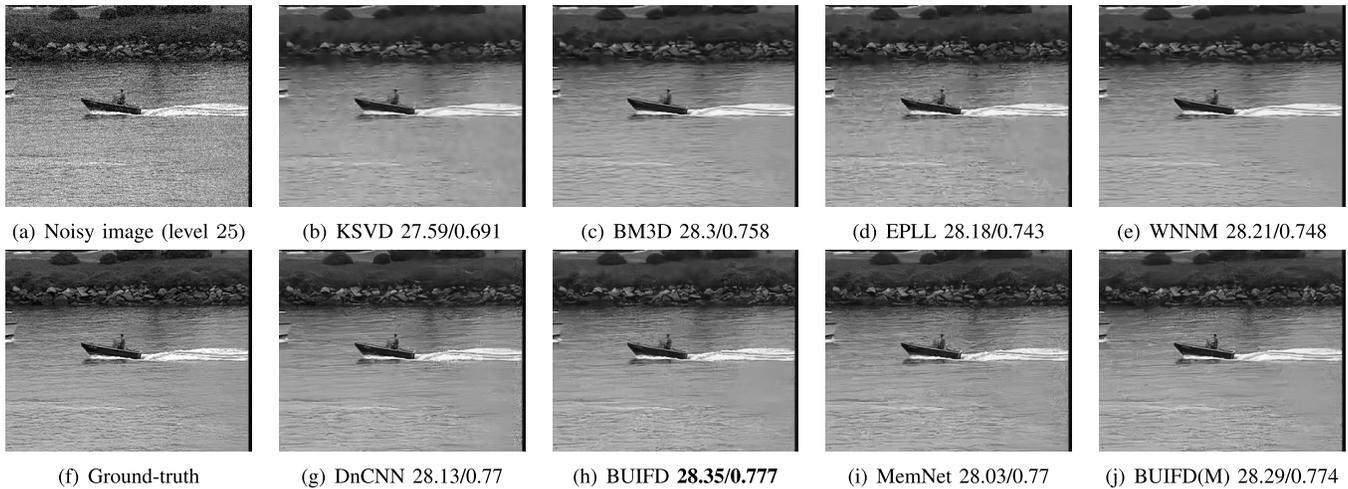

| (a) Noisy image (level 25) | (b) KSVD 27.59/0.691 | (c) BM3D 28.3/0.758 | (d) EPLL 28.18/0.743 | (e) WNNM 28.21/0.748 |
| (f) Ground-truth | (g) DnCNN 28.13/0.77 | (h) BUIFD **28.35/0.777** | (i) MemNet 28.03/0.77 | (j) BUIFD(M) 28.29/0.774 |

Fig. 8. Sample visual result from Set14, with PSNR($dB$)/SSIM values. The top row shows *non-blind* results with the traditional methods KSVD, BM3D, EPLL and WNNM, as the noise level is 25, which the default set when the noise level is unknown. And the bottom row shows the results with the different learning methods.

## V. CONCLUSION

We define a theoretical framework under which we derive an optimal denoising solution that we call fusion denoising. We integrate it into a deep learning architecture and compare with the optimal mathematical solution and with the state-of-the-art blind universal denoiser DnCNN. Our synthetic experimental results show that our Fusion Net generalizes far better to higher unseen noise levels.

We learn a data-dependent fusion function to adapt our fusion denoising network to real images. Our blind universal image fusion denoising network BUIFD improves the state-of-the-art real image denoising performance both on training noise levels and on unseen noise levels.

## REFERENCES

[1] M. Aharon, M. Elad, and A. Bruckstein, "K-SVD: An algorithm for designing overcomplete dictionaries for sparse representation," *IEEE Trans. Signal Process.*, vol. 54, no. 11, pp. 4311–4322, Nov. 2006.

[2] Y. Bengio, A. Courville, and P. Vincent, "Representation learning: A review and new perspectives," *IEEE Trans. Pattern Anal. Mach. Intell.*, vol. 35, no. 8, pp. 1798–1828, Aug. 2013.

[3] T. Brooks, B. Mildenhall, T. Xue, J. Chen, D. Sharlet, and J. T. Barron, "Unprocessing images for learned raw denoising," in *Proc. IEEE/CVF Conf. Comput. Vis. Pattern Recognit. (CVPR)*, Jun. 2019, pp. 11036–11045.

[4] A. Buades, B. Coll, and J.-M. Morel, "Nonlocal image and movie denoising," *Int. J. Comput. Vis.*, vol. 76, no. 2, pp. 123–139, Jul. 2007.

[5] H. C. Burger, C. J. Schuler, and S. Harmeling, "Image denoising: Can plain neural networks compete with BM3D?" in *Proc. CVPR*, Jun. 2012, pp. 2392–2399.

[6] X. Chen, Y. Duan, R. Houthooft, J. Schulman, I. Sutskever, and P. Abbeel, "InfoGAN: Interpretable representation learning by information maximizing generative adversarial nets," in *Proc. NeruIPS*, 2016, pp. 2172–2180.

[7] Y. Chen and T. Pock, "Trainable nonlinear reaction diffusion: A flexible framework for fast and effective image restoration," *IEEE Trans. Pattern Anal. Mach. Intell.*, vol. 39, no. 6, pp. 1256–1272, Jun. 2017.

[8] K. Dabov, A. Foi, V. Katkovnik, and K. Egiazarian, "Image denoising by sparse 3-D transform-domain collaborative filtering," *IEEE Trans. Image Process.*, vol. 16, no. 8, pp. 2080–2095, Aug. 2007.

[9] K. Dabov, A. Foi, V. Katkovnik, and K. Egiazarian, "Color image denoising via sparse 3D collaborative filtering with grouping constraint in luminance-chrominance space," in *Proc. IEEE Int. Conf. Image Process.*, Sep. 2007, pp. 313–316.

[10] W. Dong, L. Zhang, and G. Shi, "Centralized sparse representation for image restoration," in *Proc. Int. Conf. Comput. Vis.*, Nov. 2011, pp. 1259–1266.

[11] W. Dong, L. Zhang, G. Shi, and X. Li, "Nonlocally centralized sparse representation for image restoration," *IEEE Trans. Image Process.*, vol. 22, no. 4, pp. 1620–1630, Apr. 2013.

[12] M. E. Helou, S. Mandt, A. Krause, and P. Beardsley, "Mobile robotic painting of texture," in *Proc. Int. Conf. Robot. Autom. (ICRA)*, May 2019, pp. 640–647.

[13] M. E. Helou, Z. Sadeghipoor, and S. Susstrunk, "Correlation-based deblurring leveraging multispectral chromatic aberration in color and near-infrared joint acquisition," in *Proc. IEEE Int. Conf. Image Process. (ICIP)*, Sep. 2017, pp. 1402–1406.

[14] M. Elad, *Image Denoising*. New York, NY, USA: Springer, 2010, pp. 273–307.

[15] M. Elad and M. Aharon, "Image denoising via sparse and redundant representations over learned dictionaries," *IEEE Trans. Image Process.*, vol. 15, no. 12, pp. 3736–3745, Dec. 2006.

[16] C. Godard, K. Matzen, and M. Uyttendaele, "Deep burst denoising," in *Proc. ECCV*, 2018, pp. 538–554.

[17] S. Gu, L. Zhang, W. Zuo, and X. Feng, "Weighted nuclear norm minimization with application to image denoising," in *Proc. IEEE Conf. Comput. Vis. Pattern Recognit.*, Jun. 2014, pp. 2862–2869.

[18] M. Hasan and M. R. El-Sakka, "Improved BM3D image denoising using SSIM-optimized Wiener filter," *EURASIP J. Image Video Process.*, vol. 2018, no. 1, p. 25, Apr. 2018.

[19] K. He, X. Zhang, S. Ren, and J. Sun, "Deep residual learning for image recognition," in *Proc. IEEE Conf. Comput. Vis. Pattern Recognit. (CVPR)*, Jun. 2016, pp. 770–778.

[20] J.-B. Huang, A. Singh, and N. Ahuja, "Single image super-resolution from transformed self-exemplars," in *Proc. IEEE Conf. Comput. Vis. Pattern Recognit. (CVPR)*, Jun. 2015, pp. 5197–5206.

[21] S. Ioffe and C. Szegedy, "Batch normalization: Accelerating deep network training by reducing internal covariate shift," in *Proc. ICML*, 2015, pp. 448–456.

[22] D. P. Kingma and J. Ba, "Adam: A method for stochastic optimization," 2014, *arXiv:1412.6980*. [Online]. Available: http://arxiv.org/abs/1412.6980

[23] D. Krishnan and R. Fergus, "Fast image deconvolution using hyperlaplacian priors," in *Proc. NeurIPS*, 2009, pp. 1033–1041.

[24] Y. Le Montagner, E. D. Angelini, and J.-C. Olivo-Marin, "An unbiased risk estimator for image denoising in the presence of mixed Poisson–Gaussian noise," *IEEE Trans. Image Process.*, vol. 23, no. 3, pp. 1255–1268, Mar. 2014.

[25] S. Lefkimmiatis, "Non-local color image denoising with convolutional neural networks," in *Proc. IEEE Conf. Comput. Vis. Pattern Recognit. (CVPR)*, Jul. 2017, pp. 3587–3596.

[26] S. Lefkimmiatis, "Universal denoising networks : A novel CNN architecture for image denoising," in *Proc. IEEE/CVF Conf. Comput. Vis. Pattern Recognit.*, Jun. 2018, pp. 3204–3213.

[27] D. Liu, B. Wen, X. Liu, Z. Wang, and T. Huang, "When image denoising meets high-level vision tasks: A deep learning approach," in *Proc. 27th Int. Joint Conf. Artif. Intell.*, Jul. 2018, pp. 842–848.

[28] Y.-C. Liu, Y.-Y. Yeh, T.-C. Fu, S.-D. Wang, W.-C. Chiu, and Y.-C.-F. Wang, "Detach and adapt: Learning cross-domain disentangled deep representation," in *Proc. IEEE/CVF Conf. Comput. Vis. Pattern Recognit.*, Jun. 2018, pp. 8867–8876.

[29] M. A. Lobas *et al.*, "A genetically encoded single-wavelength sensor for imaging cytosolic and cell surface ATP," *Nature Commun.*, vol. 10, no. 1, p. 711, 2019.

[30] J. Mairal, F. Bach, J. Ponce, G. Sapiro, and A. Zisserman, "Non-local sparse models for image restoration," in *Proc. IEEE 12th Int. Conf. Comput. Vis.*, Sep. 2009, pp. 2272–2279.

[31] M. Makitalo and A. Foi, "Noise parameter mismatch in variance stabilization, with an application to Poisson–Gaussian noise estimation," *IEEE Trans. Image Process.*, vol. 23, no. 12, pp. 5348–5359, Dec. 2014.

[32] D. Martin *et al.*, "A database of human segmented natural images and its application to evaluating segmentation algorithms and measuring ecological statistics," in *Proc. ICCV*, 2001, pp. 416–423.

[33] Y. Matsui *et al.*, "Sketch-based manga retrieval using manga109 dataset," *Multimedia Tools Appl.*, vol. 76, no. 20, pp. 21811–21838, Nov. 2016.

[34] V. Nair and G. E. Hinton, "Rectified linear units improve restricted Boltzmann machines," in *Proc. ICML*, 2010, pp. 807–814.

[35] S. Osher, M. Burger, D. Goldfarb, J. Xu, and W. Yin, "An iterative regularization method for total variation-based image restoration," *Multiscale Model. Simul.*, vol. 4, no. 2, pp. 460–489, Jan. 2005.

[36] N. Parikh *et al.*, "Proximal algorithms," *Found. Trends Optim.*, vol. 1, no. 3, pp. 127–239, 2014.

[37] P. Perona and J. Malik, "Scale-space and edge detection using anisotropic diffusion," *IEEE Trans. Pattern Anal. Mach. Intell.*, vol. 12, no. 7, pp. 629–639, Jul. 1990.

[38] T. Plotz and S. Roth, "Benchmarking denoising algorithms with real photographs," in *Proc. IEEE Conf. Comput. Vis. Pattern Recognit. (CVPR)*, Jul. 2017, pp. 2750–2759.

[39] T. Remez, O. Litany, R. Giryes, and A. M. Bronstein, "Class-aware fully convolutional Gaussian and Poisson denoising," *IEEE Trans. Image Process.*, vol. 27, no. 11, pp. 5707–5722, Nov. 2018.

[40] S. Romdhani and T. Vetter, "Estimating 3D shape and texture using pixel intensity, edges, specular highlights, texture constraints and a prior," in *Proc. CVPR*, 2005, pp. 986–993.

[41] S. Roth and M. J. Black, "Fields of experts," *Int. J. Comput. Vis.*, vol. 82, no. 2, pp. 205–229, Jan. 2009.

[42] L. I. Rudin, S. Osher, and E. Fatemi, "Nonlinear total variation based noise removal algorithms," *Phys. D, Nonlinear Phenomena*, vol. 60, nos. 1–4, pp. 259–268, Nov. 1992.

[43] U. Schmidt and S. Roth, "Shrinkage fields for effective image restoration," in *Proc. IEEE Conf. Comput. Vis. Pattern Recognit.*, Jun. 2014, pp. 2774–2781.

[44] J.-L. Starck, E. J. Candes, and D. L. Donoho, "The curvelet transform for image denoising," *IEEE Trans. Image Process.*, vol. 11, no. 6, pp. 670–684, Jun. 2002.

[45] M. Suganuma, M. Ozay, and T. Okatani, "Exploiting the potential of standard convolutional autoencoders for image restoration by evolutionary search," in *Proc. ICML*, 2018, pp. 4778–4787.

[46] L. Sun and J. Hays, "Super-resolution from Internet-scale scene matching," in *Proc. IEEE Int. Conf. Comput. Photogr. (ICCP)*, Apr. 2012, pp. 1–12.

[47] Y. Tai, J. Yang, X. Liu, and C. Xu, "MemNet: A persistent memory network for image restoration," in *Proc. IEEE Int. Conf. Comput. Vis. (ICCV)*, Oct. 2017, pp. 4539–4547.

[48] C. Tian, Y. Xu, L. Fei, and K. Yan, "Deep learning for image denoising: A survey," 2018, *arXiv:1810.05052*. [Online]. Available: http://arxiv.org/abs/1810.05052

[49] G. Wang, C. Lopez-Molina, and B. D. Baets, "Blob reconstruction using unilateral second order Gaussian kernels with application to high-ISO long-exposure image denoising," in *Proc. IEEE Int. Conf. Comput. Vis. (ICCV)*, Oct. 2017, pp. 4817–4825.

[50] Z. Wang, A. C. Bovik, H. R. Sheikh, and E. P. Simoncelli, "Image quality assessment: From error visibility to structural similarity," *IEEE Trans. Image Process.*, vol. 13, no. 4, pp. 600–612, Apr. 2004.

[51] Y. Weiss and W. T. Freeman, "What makes a good model of natural images?" in *Proc. CVPR*, 2007, pp. 1–8.

[52] J. Xie, L. Xu, and E. Chen, "Image denoising and inpainting with deep neural networks," in *Proc. NeurIPS*, 2012, pp. 341–349.

[53] J. Xu, L. Zhang, W. Zuo, D. Zhang, and X. Feng, "Patch group based nonlocal self-similarity prior learning for image denoising," in *Proc. IEEE Int. Conf. Comput. Vis. (ICCV)*, Dec. 2015, pp. 244–252.

[54] Z. Yi, H. Zhang, P. Tan, and M. Gong, "DualGAN: Unsupervised dual learning for Image-to-Image translation," in *Proc. IEEE Int. Conf. Comput. Vis. (ICCV)*, Oct. 2017, pp. 2849–2857.

[55] Y. Jing, Y. Yang, Z. Feng, J. Ye, Y. Yu, and M. Song, "Neural style transfer: A review," 2017, *arXiv:1705.04058*. [Online]. Available: http://arxiv.org/abs/1705.04058

[56] S. Zagoruyko and N. Komodakis, "Wide residual networks," in *Proc. Brit. Mach. Vis. Conf.*, 2016, pp. 1–15.

[57] A. Zamir, A. Sax, W. Shen, L. Guibas, J. Malik, and S. Savarese, "Taskonomy: Disentangling task transfer learning," in *Proc. 28th Int. Joint Conf. Artif. Intell.*, Aug. 2019, pp. 3712–3722.

[58] J. Zhang, W. Li, and P. Ogunbona, "Joint geometrical and statistical alignment for visual domain adaptation," in *Proc. IEEE Conf. Comput. Vis. Pattern Recognit. (CVPR)*, Jul. 2017, pp. 1859–1867.

[59] K. Zhang, W. Zuo, Y. Chen, D. Meng, and L. Zhang, "Beyond a Gaussian denoiser: Residual learning of deep CNN for image denoising," *IEEE Trans. Image Process.*, vol. 26, no. 7, pp. 3142–3155, Jul. 2017.

[60] K. Zhang, W. Zuo, S. Gu, and L. Zhang, "Learning deep CNN denoiser prior for image restoration," in *Proc. CVPR*, 2017, pp. 3929–3938.

[61] K. Zhang, W. Zuo, and L. Zhang, "FFDNet: Toward a fast and flexible solution for CNN-based image denoising," *IEEE Trans. Image Process.*, vol. 27, no. 9, pp. 4608–4622, Sep. 2018.

[62] K. Zhang, W. Zuo, and L. Zhang, "Learning a single convolutional super-resolution network for multiple degradations," in *Proc. IEEE/CVF Conf. Comput. Vis. Pattern Recognit.*, Jun. 2018, pp. 3262–3271.

[63] Y. Zhang, Y. Guo, Y. Jin, Y. Luo, Z. He, and H. Lee, "Unsupervised discovery of object landmarks as structural representations," in *Proc. IEEE/CVF Conf. Comput. Vis. Pattern Recognit.*, Jun. 2018, pp. 2694–2703.

[64] D. Zoran and Y. Weiss, "From learning models of natural image patches to whole image restoration," in *Proc. Int. Conf. Comput. Vis.*, Nov. 2011, pp. 479–486.